# Machine Learning-Based Vehicle Intention Trajectory Recognition and Prediction for Autonomous Driving

Hanyi Yu1*
Computer Science
University of Southern California
Los Angeles, CA, USA
* Corresponding author: e-mail:hanyiyu@usc.edu

Shuning Huo2
Statistics
Virginia Tech
Blacksburg, VA, USA
shuni93@vt.edu

Mengran Zhu3
Computer Engineering
Miami University
Oxford, OH USA
mengran.zhu0504@gmail.com

Yulu Gong4
Computer & Information Technology
Northern Arizona University
Flagstaff, AZ, USA
yg486@nau.edu

Yafei Xiang5
Computer Science
Northeastern University
Boston, MA, USA
xiang.yaf@northeastern.edu

*Abstract*—In recent years, the expansion of internet technology and advancements in automation have brought significant attention to autonomous driving technology. Major automobile manufacturers, including Volvo, Mercedes-Benz, and Tesla, have progressively introduced products ranging from assisted-driving vehicles to semi-autonomous vehicles. However, this period has also witnessed several traffic safety incidents involving self-driving vehicles. For instance, in March 2016, a Google self-driving car was involved in a minor collision with a bus. At the time of the accident, the autonomous vehicle was attempting to merge into the right lane but failed to dynamically respond to the real-time environmental information during the lane change. It incorrectly assumed that the approaching bus would slow down to avoid it, leading to a low-speed collision with the bus. This incident highlights the current technological shortcomings and safety concerns associated with autonomous lane-changing behavior, despite the rapid advancements in autonomous driving technology. Lane-changing is among the most common and hazardous behaviors in highway driving, significantly impacting traffic safety and flow. Therefore, lane-changing is crucial for traffic safety, and accurately predicting drivers' lane change intentions can markedly enhance driving safety. This paper introduces a deep learning-based prediction method for autonomous driving lane change behavior, aiming to facilitate safe lane changes and thereby improve road safety.

*Keywords-Autonomous driving; CNN-LSTM model; Deep learning; Autonomous lane change*

I. INTRODUCTION

The purpose of autonomous lane change is to achieve faster driving speed or better driving conditions. Most existing research on the modeling of autonomous lane change decisions has ignored the human factor, which is very important for the accurate modeling of autonomous lane change decision. In this paper, a new autonomous lane change decision model is proposed by integrating the human factors represented by driving style. The model takes into account not only the surrounding traffic information, but also the driving style of the surrounding vehicles to make lane change/lane keeping decisions. In addition, the model can mimic the decision-making operations of a human driver by learning the driving style of a self-driving car. The results show that the model can accurately describe the human decision-making strategy, simulate the human driver's lane change action, and the prediction accuracy can reach 98.66%.

In traditional mathematical lane change decision-making models, some of them focus on describing the overall impact of lane change on traffic flow, ignoring the consideration of human drivers. Except for the lane change decision model based on game theory, most models treat the lane change decision process as a single driver decision, and do not capture the interaction between lane change vehicles and surrounding vehicles. The rules and parameters defined in most existing models are limited and few in number, which cannot accurately describe lane change behavior. The lane change decision time is often incorrectly labeled as crossing time. The above problems lead to the poor prediction accuracy of the traditional lane change decision model. To overcome the limitations of traditional lane change decision models, an autonomous lane change decision model (DSA-DLC) that considers driving style perception was proposed. The model implicitly represents the driving style of vehicles by extracting it from the driving operation graph (DOP) of historical trajectories, and the driving style of surrounding vehicles and autonomous vehicles was taken as human factors. The relationship between lane change decision, traffic factor and human factor is modeled by neural network, and the autonomous lane change decision model is constructed to predict the autonomous lane change decision of human driver.

To solve the above problems, this paper proposes a CNN-LSTM lane change intention prediction model based on deep learning, which combines the advantages of convolutional neural network (CNN) and long short-term memory network (LSTM) to improve road traffic safety and fluency. The model first processes high-dimensional spatiotemporal traffic data through CNN to extract key features, including vehicle speed, acceleration, relative position and other information. Subsequently, the LSTM uses these features to grasp the time-series nature of traffic flow and capture the long-term dependencies in driving behavior. This combination enables the model to effectively predict the driver's lane change intention and provide more accurate decision support for intelligent transportation systems and autonomous driving technologies. On this basis, future studies can further explore the adaptability of the model in different traffic scenarios and how to optimize the model performance through real-time data feedback.

## II. RELEVANT THEORIES AND MODELS

### A. Behavioral prediction in autonomous driving systems

The behavior prediction in the automatic driving system is to accurately detect the current position, speed, direction of movement and other information of the surrounding vehicles through various sensors installed by the autonomous vehicle itself, and then use it to predict its future trajectory. In recent years, vehicle trajectory prediction methods are mainly divided into two categories: reinforcement learning model based method and neural network-based interactive learning and social perception method. Model-based methods use appropriate learning algorithms to achieve results in specific scenarios. In recent years, the method based on the game model has often been used for trajectory prediction in intelligent vehicles. Li et al. The CMMetric algorithm used for behavior modeling and prediction is used to sort the CMetric values of the surrounding vehicles to determine the passing order of the surrounding vehicles and predict the trajectory. The above two methods based on the game model are aimed at the game between the two players, and can not meet the trajectory prediction of the target vehicle in many scenarios.

### B. Machine learning predictive models

The success of deep learning in many practical applications has prompted research into its application in motion prediction. With the recent success of recurrent neural networks (RNNS), one line of research called short-time memory (LSTM) has been used for sequence prediction tasks. The authors used LSTM to predict the future trajectory of pedestrians in social interactions. In the literature, LSTM has been applied to predict vehicle position using past trajectory data. Another RNN variant, called a gated recursive unit (GRU), is combined with conditional variational autoencoders (CVAE) to predict vehicle trajectory. In addition, by applying convolutional neural networks (CNNS) to a series of visual images, the motion of a simple physical system is predicted directly from image pixels. Later, researchers proposed a system in which CNNs were used to predict short-term vehicle trajectories, with BEV raster images encoding the surroundings of a single participant as input, which were then also applied to vulnerable traffic participants.Despite the success of these methods, they do not address the underlying multimodal problem of possible future trajectories required for accurate long-term traffic prediction.

At present, there are many researches to solve the problem of multimodal modeling. Mixed density networks (MDNs) are traditional neural networks that solve multimodal regression problems by learning the parameters of Gaussian mixture models. However, MDNs is often difficult to train in practice due to numerical instability when operating in high-dimensional Spaces. Therefore, based on the above practical problems, the method proposed in this paper directly calculates the prediction results of multiple modes on a single forward CNN model.

### C. Construction of lane change decision model

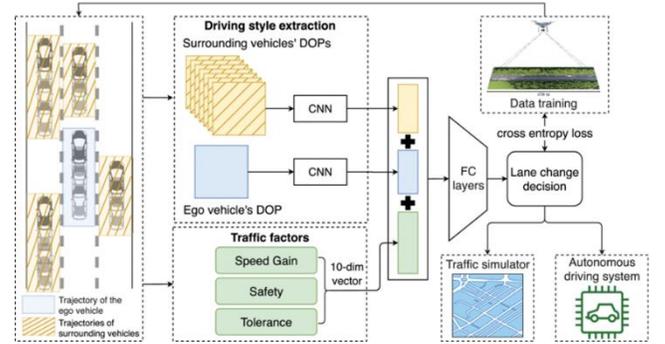

Figure 1. DSA-DLC decision model architecture

y is the lane change decision vector, which includes three elements corresponding to the probabilities of lane keeping, lane changing left and lane changing right respectively. Considering the lane change incentive factor, safety factor, tolerance factor and driving style extracted from DOP D, the decision-making formula is as follows:

$$y = f_{LC}(F_{incentive}, F_{SAFETY}, F_{tolerance}, F_{DS}(D)) \quad (1)$$

Hypothesis 1:
$$v_i, \ d_i, \ i \in \{E, P, PL, PR, FL, FR, ASL, ASR\} \quad (2)$$

Where v is the speed of vehicle i and d is the longitudinal distance from vehicle i to vehicle I.

- Lane change facilitation factor

$$F_{incentive} = F_I((v_E - v_P), (v_{PL} - v_P), (v_{PR} - v_P), (d_{PL} - d_P), (d_{PR} - d_P)) \quad (3)$$

- (2) Lane change safety factor

$$F_{safety} = f_s(d_{FL}, d_{FR}, (v_E - v_{FL}), (v_E - v_{FR})) \quad (4)$$

The risk of collision is affected by the distance and speed difference between the vehicle and the vehicle behind the target lane. When the distance between the self vehicle and the rear vehicle is large enough, and the speed of the self vehicle is higher than the speed of the rear vehicle in the target lane, the collision risk is low.

- (3) Tolerance factor

$$F_{tolerance} = f_T(d_P - v_E \cdot t_h) \quad (5)$$

It is used to measure the driving condition in the current lane and is highly correlated with the safe headway of the car and the distance of the car in front. Depending on driving experience, drivers may be more inclined to stay farther than a safe distance on the highway to improve driving comfort.

Among them, f_I, f_s and f_T are constructed by neural network as traffic factors input to the lane change decision model.

- (4) Model design

$$y = f_{LC}(F_{incentive}, F_{SAFETY}, F_{tolerance}, F_{DS}(D)) =$$
$$f_\theta \begin{pmatrix} v_E - v_P, \ v_{PL} - v_P, \ v_{PR} - v_P, \\ d_{PL} - d_P, \ d_{PR} - d_P, \ d_{FL}, \ d_{FR}, \ v_E - v_{FL}, \\ v_E - v_{FR}, \ d_P - v_E \cdot t_h, \ D \end{pmatrix}$$
(6)

theta is a set of parameters of the model, and the lane change decision is a multi-parameter nonlinear problem, which is solved by deep learning. Using trajectory data set, input variables and decisions are extracted frame-by-frame and modeled as supervised learning problems. The learning goal is to find a model to minimize the long-term average loss, calculate the loss value and optimize the neural network, defined as:

$$\frac{1}{N}\sum_{i=1}^{N} L(\hat{y}_i, f_\theta(.)) = \frac{(-1)}{N}\sum_{i=1}^{N}\sum_{c=1}^{3} \hat{y}_{ic} \log(y_{ic}) \quad (7)$$

The DSA-DLC decision model consists of convolutional neural network (CNN) and fully connected neural network (FC). CNN has shown a strong image classification ability in capturing the hidden features of images. The driving style is regarded as the hidden characteristics of the vehicle, and CNNs is used to extract the hidden driving style from the DOP. Since the driving styles of self-driving cars and surrounding cars have different effects on lane change decisions, two CNNs are used, one for the self-driving car DOP and the other for the surrounding car DOP.

D. *Neural networks and attention mechanisms*

Model-based methods include Gauss process model and social force model. Minguez et al. proposed a method to predict pedestrian trajectory by collecting the information of key parts of the pedestrian body, such as the shoulder, and applying the Gauss process dynamics model based on balance. Based on the social force model, Rinke et al. proposed a multi-level description method of road user movement and its interaction, and discussed pedestrian movement target points and possible trajectories in layers. By first determining the target point, using Lagrange polynomial to estimate other trajectories in turn, and then using the conflict avoidance strategy based on social force model to select trajectories, the best predicted trajectories are generated.

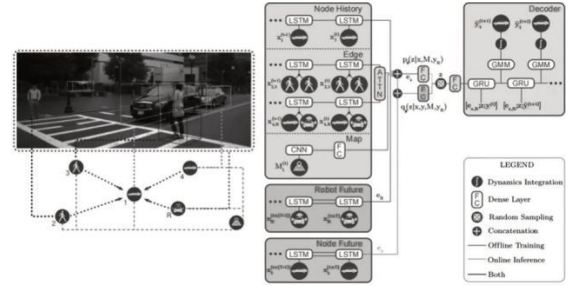

Figure 2. Framework of trajectory prediction network for trajectory ++

However, model-based methods are highly dependent on theories and scenarios, and require a large amount of data to train models. Therefore, in recent years, more researchers are still committed to methods based on the combination of neural networks and attention mechanisms, generally RNN/LSTM/GRU and other recurrent neural networks and their variants to combine attention mechanisms. Zhou et al. use recurrent neural Networks (RNN) and Graph Convolutional Networks (GCN) to simulate the state of pedestrians and their interactions by considering the movement information of individual pedestrians and their interactions with surrounding pedestrians (Figure 3). The trajectory is predicted by combining attention mechanism and companion loss function.

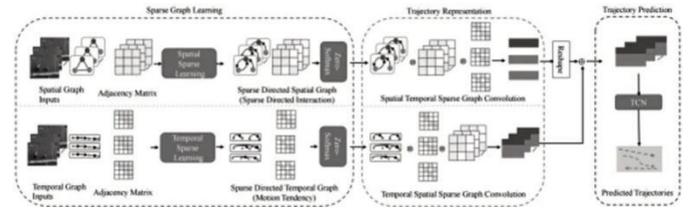

Figure 3. SGCN network framework

Liu et al. proposed a prediction model of whether pedestrians cross the road based on graph convolution, and predicted pedestrian trajectory within a future time range by reasoning only about the relationship between pedestrians and the surrounding environment and their own body movements. Shi et al. proposed a novel analytic Graph Convolution Network (SGCN), which combined Sparse directed interaction with motion trends for pedestrian trajectory prediction. The model used a sparse graph learning method, based on an attention mechanism, to score track points. It is fed back into the asymmetric convolutional network to obtain high-level interaction features. The adjacency matrix obtained after normalization can represent the sparse graph. Finally, the trajectory is predicted by combining the parameters of double Gaussian distribution estimated by the graph convolutional network for trajectory prediction. Compared with previous predictive modeling methods, it makes targeted selection of interactive pedestrian information, vehicle information and environmental information, instead of directly applying all the above information within a certain range to modeling without difference.

In conclusion, The main advantage of the method based on deep learning neural network and long short-term memory network (LSTM) to predict lane change behavior of autonomous driving is that it can effectively process and analyze time series data, so as to accurately predict lane change behavior of vehicles in different traffic situations. By learning from historical data, this method predicts future driving paths and potential risks in real time, supporting autonomous driving systems. It plays a crucial role in enhancing the safety and efficiency of autonomous vehicles. By predicting the potential behavior of surrounding vehicles, the autonomous driving system can make adjustments in advance or take avoidance measures to avoid traffic accidents and ensure the safety of passengers and pedestrians. In addition, this method helps to improve the smoothness of traffic flow, reduce unnecessary lane changes and braking, and thus improve road utilization and driving efficiency.

## III. METHODOLOGY

### A. Data extraction and processing

- Data set

The data set in this paper is primarily derived from the Federal Highway Administration's Next Generation Simulation (NGSIM) dataset for extracting vehicle trajectories and modeling lane change predictions, which has been adopted by many previous studies. In a time interval of 0:1 second, the dataset recorded the position, speed, acceleration, and headway information of each vehicle on U.S. highways 101 and Interstate 80 (I-80). Both locations contain 45 minutes of vehicle trajectory data. Highway 101 is 640 meters long and has five main lanes and six service lanes, while I-80 is about 500 meters long and has six main lanes.

In this paper, six vehicle trajectory data sequences were extracted from NGSIM, each of which lasted 10 minutes. We removed the first 5 minutes from each 15-minute sequence to ensure a sufficient number of vehicles in each frame. For each sequence, the first 2 minutes are defined as the test set and the remaining 8 minutes are defined as the training set. Since the data is recorded at 10 frames per second, we can get a total of 1,200 test time steps and 4,800 training time steps.

- Data tag

Vehicles are marked as "intending to change lanes to the left," "intending to travel along the lane," or "intending to change lanes at each time step." The way we mark the status of the vehicle is as follows.

As shown in Figure 4, we first collect all lane exchange points, i.e. the point at which the vehicle gravity point crosses the dotted line dividing the lane, the vehicle. If the vehicle is at the lane change point at time step t, we check its trajectory at [t-δt, t+δt] (δt=2s) and calculate its heading value θ for that time period. We mark the start and end of this lane change trajectory when θ reaches the boundary value θ bound :| θ |= θ bound.

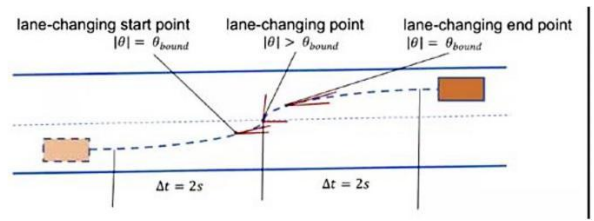

Figure 4. The start, change point and end point of the lane change trajectory

Figure 5.(a) The lane change prediction point is determined if the vehicle is predicted to enter the lane change for 3 consecutive time steps. The lane change prediction time is defined as the time interval between the lane change point and the lane change prediction point. (b) n continuous time steps are packaged into a trajectory segment. If the NTH time step of the track segment is the lane following time step, then the segment is a lane following segment, otherwise it is marked as a lane changing segment.

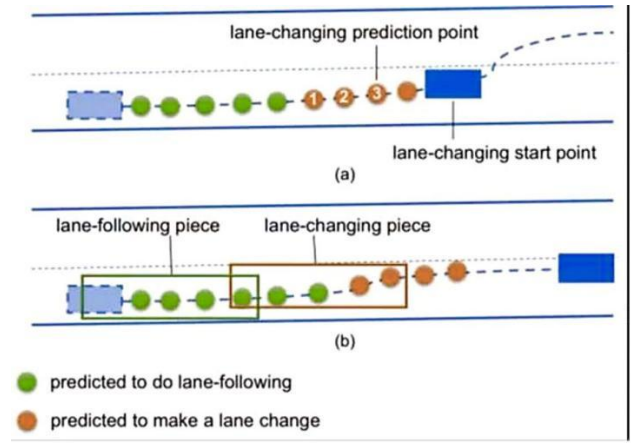

Figure 5. Vehicle behavior marks track segments

Figure 5 (b) depicts the way we label track segments. For each vehicle, n consecutive time steps are packaged into a track segment. If the NTH time step of the track segment is the lane change time step, then the segment is a lane change segment, otherwise it is marked as a lane following segment. In this article, we set n to 6,9, and 12 to determine the effect of the length of the historical track on the final result.

In summary, this paper can obtain about 60,000 lane changes, plus 400,000 cars for training. This obviously involves a problem of data imbalance, in which there are many more lane following pieces than lane changing pieces used for training, which will lead to overfitting during training. To solve this problem, we randomly selected the same number of fragments N from the change-left pool, the change-after pool, and the change-right pool and mixed them together as the training data set. To maximize the use of data, N is set to the number of pieces (30,000 pieces) in the pool on the right side of the lane change.

Then, given the first (n1) time step historical track and neighbor information in the test set, the lane change intent of each vehicle is predicted at each time step. The predicted time of lane change is also calculated after filtering the results. Specifically, the lane change prediction point is determined if the forecast vehicle makes 3 continuous time steps of lane change, and the lane change prediction time is defined as the time interval between the lane change point and the lane - the change prediction point, as shown in Figure 5 (a).

*B. Method*

In this paper you need to predict whether the car will change lanes and which lane it will merge into. We use an LSTM to enable the agent to reason about the vehicle's historical trajectory information. However, since human decision-making behavior will also depend on the surrounding vehicles, we also use vehicle neighborhood information as input to the network.

- Input function

Two types of input features are used for the prediction algorithm:

(1) Information about the vehicle itself and (2) information about the vehicle's neighbors. Information about the vehicle itself includes:

a) Vehicle acceleration

b) The steering Angle of the vehicle relative to the road

c) Global lateral vehicle position relative to the lane

d) Global longitudinal vehicle position relative to the lane

Vehicles estimating their lane change intentions are provided by the following features:

e) Presence of the left lane (1 if present, 0 otherwise)

f) Presence of the right lane (1 if present, 0 otherwise)

g) Longitudinal distance between self vehicle and left front vehicle

h) The longitudinal distance between the self vehicle and the vehicle in front

i) Longitudinal distance between self vehicle and right front vehicle

j) Longitudinal distance between self vehicle and left rear vehicle

k) Longitudinal distance between self vehicle and rear vehicle

l) Longitudinal distance between self vehicle and right rear vehicle

- Network structure

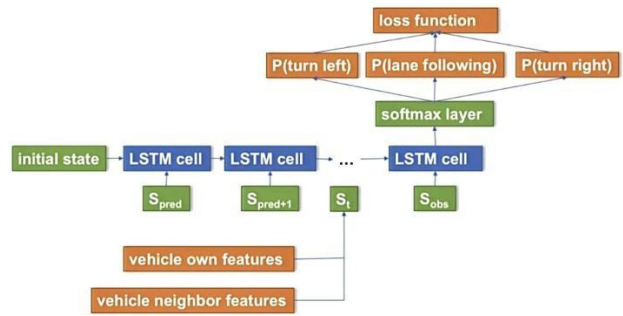

Figure 6.  LSTM network structure for lane change intent prediction

As shown in Figure 6, we adopt the LSTM network architecture to deal with this problem of intention prediction of lane change. The embedded dimension selected for the features of the vehicle itself and its neighbors is 64, and the hidden dimension of the LSTM network is 128. We chose a learning rate of 0:00 0125 and used the soft-max cross-entropy loss as the training loss: loss =- Σ i=1yi´log (yi). Where y is the true label of the intention of the i lane change (yi´=1, intention exists, yi´=0, intention does not exist. i∈ {1,2,3}. y1´ is the left intent to change lanes, y2´ is the intent to follow lanes, and y3´ is the right intent to change lanes). yi is the predicted output probability of the model with the i lane change intention after passing through the soft-max layer.

TABLE I.  COMPARISON OF THE ACCURACY OF CHANGING LANE PREDICTIONS

|  | Real Predict | Left | Following | Right |
|---|---|---|---|---|
| SA-LSTM | Left | 87.40% | 12.34% | 0.26% |
|  | Following | 7.47% | 85.33% | 7.20% |
|  | Right | 2.94% | 11.22% | 85.84% |
| Feedforward Neural Network | Left | 84.6% | 15.40% | 0% |
|  | Following | 2.61% | 83.78% | 13.61% |
|  | Right | 2.44% | 12.91% | 79.65% |
| Logistic Regression | Left | 64.91% | 35.03% | 0.06% |
|  | Following | 9.88% | 82.87% | 7.25% |
|  | Right | 0.05% | 36.30% | 63.65% |

*C. Test result*

Comparisons with other network structures are made with feedforward neural networks, logistic regression, and LSTMS without adjacent feature inputs to show the advantages of adding historical tracks and environmental factors. Table I and Figure 6 show the classification accuracy calculated by our algorithm, feedforward neural network and logistic regression. The method we call Environment Aware (SA) -LSTM, based on the advantages of historical track information and neighbor information, outperforms the other two methods in terms of prediction accuracy on all classification types (left transition track, right transition track, and right transition track).

*D. Comparison of different trajectory lengths*

Set the historical track lengths of the LSTM structure to 6,9, and 12, and compare them to each other. The results are shown in Table II and visualized in Figure 7. We compared the results of five different trajectory sequences to help us get a general idea of the trend of curve change. In all prediction scenarios, the prediction accuracy increases as the history length increases (left lane change, right lane change).

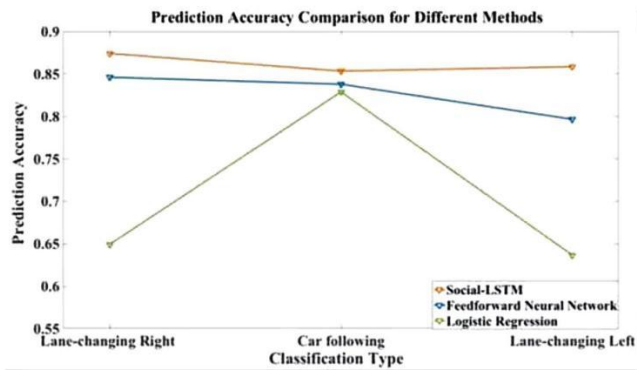

Figure 7.  Comparison of prediction accuracy of different methods

Based on the accuracy of the different methods, A-LSTM outperformed the other two types in all classification types, including right lane change, lane follow, and left lane change. Therefore, it can be seen that the longer the historical track length, the more information we can obtain from the previous track, and the higher the accuracy in the final result. However, there is a trade-off between length and computation time. As the length increases, the improvement in accuracy slows down. From the figure, we can see that length = 12 achieves only slightly higher accuracy compared to length = 9. More importantly, the length of the historical trajectory should not be too long.

## IV.  CONCLUSION

The paper presented an innovative approach to enhancing the safety and efficiency of autonomous driving through a deep learning-based prediction model for lane-changing behavior. It underscores the importance of incorporating human factors, represented by driving styles, into autonomous lane change decision models. Traditional models often overlook the interaction between vehicles and the dynamic nature of driving behaviors, leading to inaccuracies in predicting lane change decisions. By integrating CNN and LSTM networks, the model successfully captures the complex dynamics of traffic flow and driver behavior, offering a more accurate and reliable prediction mechanism. This approach not only addresses the limitations of previous models but also sets a new standard for research in autonomous driving technologies.

### A.  Experimental Conclusion Analysis

The experimental results demonstrate the model's superior performance in predicting lane change intentions with a remarkable accuracy rate of 98.66%. The use of CNN-LSTM architecture allows for the effective processing of spatiotemporal traffic data, capturing the nuanced patterns of driving behavior over time. The comparison with other network structures, such as feedforward neural networks and logistic regression, further validates the model's effectiveness. By incorporating the driving style and surrounding traffic information, the model provides a comprehensive understanding of lane change behavior, significantly outperforming traditional approaches in prediction accuracy.

### B.  Future Prospects of Deep Learning in Autonomous Driving and Beyond

The success of the CNN-LSTM model in predicting lane change intentions opens up new avenues for the application of deep learning in autonomous driving and other fields. The ability to process and analyze large datasets to predict complex behaviors has far-reaching implications, from improving road safety to optimizing traffic flow and reducing congestion. Future research can explore the adaptability of this model in different traffic scenarios and its integration into real-time autonomous driving systems. Beyond autonomous driving, the principles and methodologies developed in this study can be applied to other domains where predicting human behavior is crucial, such as robotics, smart cities, and personalized services. The convergence of deep learning, big data, and computational power promises to revolutionize how we understand and interact with the world around us, driving innovation and improving quality of life across various sectors.